# An Artificial Intelligence-Based Framework for Predicting Emergency Department Overcrowding: Development and Evaluation Study

Orhun Vural[1], MSc; Bunyamin Ozaydin[2,3], PhD; Khalid Y. Aram[4], PhD; James Booth[5], MD; Brittany F. Lindsey[6], MPA; Abdulaziz Ahmed[2,3]*, PhD

[1]Department of Electrical and Computer Engineering, University of Alabama at Birmingham, Birmingham, AL, United States

[2]Department of Health Services Administration, School of Health Professions, University of Alabama at Birmingham, Birmingham, AL, United States

[3]Department of Biomedical Informatics and Data Science, Heersink School of Medicine, University of Alabama at Birmingham, Birmingham, AL, United States

[4]School of Business & Technology, Emporia State University, Emporia, KS, United States

[5]Department of Emergency Medicine, University of Alabama at Birmingham, Birmingham, AL, United States

[6]Department of Patient Throughput, University of Alabama at Birmingham, Birmingham, AL, United States

* Corresponding Author, Abdulaziz Ahmed, PhD, Department of Health Services Administration, School of Health Professions, University of Alabama at Birmingham, Birmingham, AL 35233, United States, Email: aahmed2@uab.edu

## Abstract

**Background:** Emergency department (ED) overcrowding remains a major challenge, causing delays in care and increased operational strain. Hospital management often reacts to congestion after it occurs. Machine learning predictive modeling offers a proactive approach by forecasting patient flow metrics, such as waiting count, to improve resource planning and hospital efficiency.

**Objective:** This study develops machine learning models to predict ED waiting room occupancy at two time scales. The hourly model forecasts the waiting count six hours ahead (e.g., a 1 PM prediction for 7 PM), while the daily model estimates the average waiting count for the next 24 hours (e.g., a 5 PM prediction for the following day's average). These tools support staffing decisions and enable earlier interventions to reduce overcrowding.

**Methods:** Data from a partner hospital's ED in the southeastern United States were used, integrating internal metrics and external features. Eleven machine learning algorithms, including traditional and deep learning models, were trained and evaluated. Feature combinations were optimized, and performance was assessed across varying patient volumes and hours.

**Results:** TSiTPlus achieved the best hourly prediction (MAE: 4.19, MSE: 29.32). The mean hourly waiting count was 18.11, with a standard deviation of 9.77. Accuracy varied by hour, with MAEs ranging from 2.45 (11 PM) to 5.45 (8 PM). Extreme case analysis at one, two, and three standard deviations above the mean showed MAEs of 6.16, 10.16, and 15.59, respectively. For daily predictions, XCMPlus performed best (MAE: 2.00, MSE: 6.64), with a daily mean of 18.11 and standard deviation of 4.51.

**Conclusions:** These models accurately forecast ED waiting room occupancy and support proactive resource allocation. Their implementation has the potential to improve patient flow and reduce overcrowding in emergency care settings.

## Introduction

### Background

Emergency departments (EDs) are responsible for the majority of hospital admissions, even though most ED visits result in a discharge [1]. In 2021, there were more than 140 million visits to American EDs [2], of which 14.5 million (10.4%) led to hospital inpatient admissions, and 2 million (1.4%) led to admission to critical care units [3]. Several factors lead to ED overcrowding including complexity of complaints and injuries [4, 5], resource limitations [6], large patient volumes [7], and inefficient flow of patients [8]. ED overcrowding is associated with poor healthcare outcomes. It can lead to delays in diagnosis and treatment, which can result in poorer patient outcomes, higher comorbidities, and increased patient illness [9, 10]. For instance, it was found that patients with acute coronary syndrome who presented to overcrowded EDs had a significantly higher rate of serious complications than those who presented to a non-crowded ED (6% vs 3%). The complications include death, late myocardial infarction, cardiac arrest, arrhythmias, heart failure, stroke, and hypotension [11]. Treatment delays also lead to serious complications, including death [12]. Sprivulis et al. observed a significant linear relationship between ED overcrowding and patient mortality based on three years of data from three large hospital systems [13].

The Emergency Medicine Practice Committee of the American College of Emergency Physicians (ACEP) reported that ED overcrowding is a hospital-wide patient flow problem rather than an isolated ED problem [1, 2]. A key approach suggested by the ACEP to improve patient flow is the full capacity protocol (FCP), an internationally recognized approach used as a communication tool between ED and inpatient units [14]. FCP is a key approach to improve patient flow across the entire hospital and is a communication tool between ED and inpatient units. It contains a set of interventions that can be tailored to the severity levels of ED overcrowding [15]. A set of criteria triggers each intervention level. The criteria are based on different patient flow measures (PFMs), such as the number of patients waiting to be admitted to an inpatient unit (boarded patients) [15, 16].

In the current FCP practice, the unit that manages patient flow relies on near real-time PFM values to activate different FCP levels, which is a reactive approach. At our partner hospital, an academic medical center in the southeastern U.S., this unit is called the patient flow coordination team (PFCT). Implementing FCP interventions (e.g., creating hallway treatment spaces or activating on-call personnel) requires preparation time, forcing PFCT to prepare and act simultaneously when the ED is already overcrowded, significantly increasing their stress. Therefore, accessing the predicted PFM values before overcrowding can provide the PFCT enough time to prepare before they implement FCP interventions. Given the critical role of timely and accurate information in driving FCP interventions, it is essential to develop prediction models to forecast PFMs used by the FCP criteria, transforming FCP from reactive to proactive. Proactive FCP can help in planning and implementing interventions at different crowding severity levels proactively before the ED is already overcrowded. Predicted PFMs built into a proactive FCP can help the PFCT anticipate future FCP level escalation and prepare for interventions, such as coordinating staffing needs and creating additional hallway treatment spaces.

There are many PFMs that can be used to determine the FCP level such as ED hourly waiting count, ED boarding count, and the number of patients by Emergency Severity Index (ESI), among others. For this study, the advisory board of the project recommended that we start with building a prediction model for the ED waiting count as it is one the most important PFMs. The advisory board members represent different ED units at the partner hospital. The members include the chair of Emergency Medicine, the chief medical information officer, the associate principals of the Office of Clinical Practice Transformation (CPT), the associate vice president of clinical operations at the center of patient flow, the senior director of Emergency Services, and the nurse director for Emergency Services. Based on the feedback from the advisory board, the goal of this paper is to build deep learning models to predict the ED waiting count at two time points: 1) In the hourly basis, in which the prediction models forecast the waiting count in the next six hours; 2) In daily basis where the models predict the average waiting count for next working day. The prediction information allows PFCT to prepare ED resources to improve patient flow and consequently mitigate ED overcrowding. This study is part of a larger funded ED improvement project and the models presented in this study will be integrated in a decision support system to be used by our partner hospital.

**Prior Work**

While FCP provides a way to tackle ED overcrowding, predictive modeling aims to anticipate overcrowding events before they occur, shifting FCP from a reactive to a proactive approach. Various studies have been done in the literature to predict different ED PFMs or outcomes. Various traditional models have been developed to predict hospital admissions at ED triage, relying on a limited set of demographics, administrative, and clinical variables [17-19]. For example, Parker et al. [18] used variables such as age group, race, postal code, day of the week, time of day, triage category, mode of arrival, and fever status in a logistic regression model to predict hospital admissions with an area under the curve (AUC) of 0.825. Similarly, Sun et al. [19] developed a model to predict immediate hospital admission at the time of ED triage using routine administrative data, including age, patient acuity category, arrival mode, and coexisting chronic diseases like diabetes, hypertension, and dyslipidemia, achieving an AUC of 0.849. However, because these models focus on only individual patient-level triage outcome predictions, they offer limited utility for administrative management in addressing ED crowding, as they do not provide a comprehensive view of overall patient flow or resource utilization.

In one of the earlier studies that focused on predicting ED PFMs at the aggregate patient level, Schweigler et al.[20] developed a baseline model to predict ED overcrowding using historical averages and compared its performance to more advanced time series models, including seasonal autoregressive integrated moving average (ARIMA) and a sinusoidal model with an autoregressive error term. However, these models relied solely on historical averages to predict ED crowding based on bed occupancy, without considering any other predictors, limiting their utility for comprehensive ED crowding management. A subsequent time series modeling study by Kadri et al. [21] also developed an ARIMA model to predict daily patient attendances at a pediatric ED, relying solely on historical attendance data without incorporating external predictors.

Recent studies have leveraged advanced machine learning (ML) models to predict ED PFMs such as the number of arriving patients at hourly and/or daily levels. Harrou et al.[22] developed deep learning models to predict hourly and daily ED visits. They compared the performance of Variational AutoEncoder (VAE) with seven deep learning models, including Recurrent Neural Networks (RNN), Long Short-Term Memory (LSTM), Bidirectional LSTM (BiLSTM), Convolutional LSTM, Restricted Boltzmann Machine (RBM), Gated Recurrent Units (GRUs), and Convolutional Neural Networks (CNN). Tuominen et al.[23] used advanced ML models, including DeepAR, N-BEATS, Temporal Fusion Transformer (TFT), and LightGBM, with multivariable inputs such as bed availability in catchment area hospitals, traffic data, and weather variables to predict daily ED occupancy. Similarly, Giunta et al. [24] developed a multivariable predictive model based on the National Emergency Department Overcrowding Study (NEDOCS) score to predict sustained critical ED overcrowding lasting 8 or more hours, incorporating weather, patient flow, and bed occupancy variables. However, despite their advanced methodologies and multivariable approaches, none of these studies examined their predictions within the context of a proactive FCP.

Although the studies presented in this section represent important advancements in ED overcrowding prediction, they each have critical limitations that reduce their practical applicability in high-volume ED settings. Many models have relied on a narrow set of inputs, such as patient arrivals or historical bed occupancy, without incorporating a broader range of operational, staffing, or environmental variables. Additionally, most studies have focused exclusively on daily metrics, failing to capture both hourly and daily fluctuations needed for real-time resource planning. To address these gaps, this study integrates both hourly and daily predictions, providing a more granular and actionable forecasting tool. It also expands the range of predictor variables by evaluating different combinations across 15 distinct datasets, using multiple state-of-the-art time-series deep learning algorithms to enhance prediction accuracy. This study focuses on waiting count as the primary prediction target, marking the first in a series of PFMs for which our team will develop predictive models. The contributions of this study are as follows:

- We proposed an improvement for the current reactive FCP to make it proactive based on prediction models.
- This study is based on a real-world dataset and conducted in collaboration with key decision-makers from the ED.
- We introduced two complementary prediction approaches—hourly and daily—that provide actionable insights at different time scales.
- We developed predictive models that integrate multiple predictors from various sources to enhance accuracy and provide a comprehensive understanding of patient flow dynamics.

## Methods

### Research Framework

Figure 1 illustrates the proposed framework, which consists of three main phases: data preparation, training, and evaluation. The framework incorporates two distinct prediction approaches: hourly predictions, which forecast patient counts in the waiting room six hours ahead to support resource management throughout the day, and daily predictions, which estimate the average waiting count in the waiting room for the next 24 hours, providing insights into overall daily trends. Throughout this paper, patient counts in the waiting room are termed "waiting counts" and the average patient count in the waiting room is designated as "average waiting count."

**Figure 1.** Proposed Research Framework.

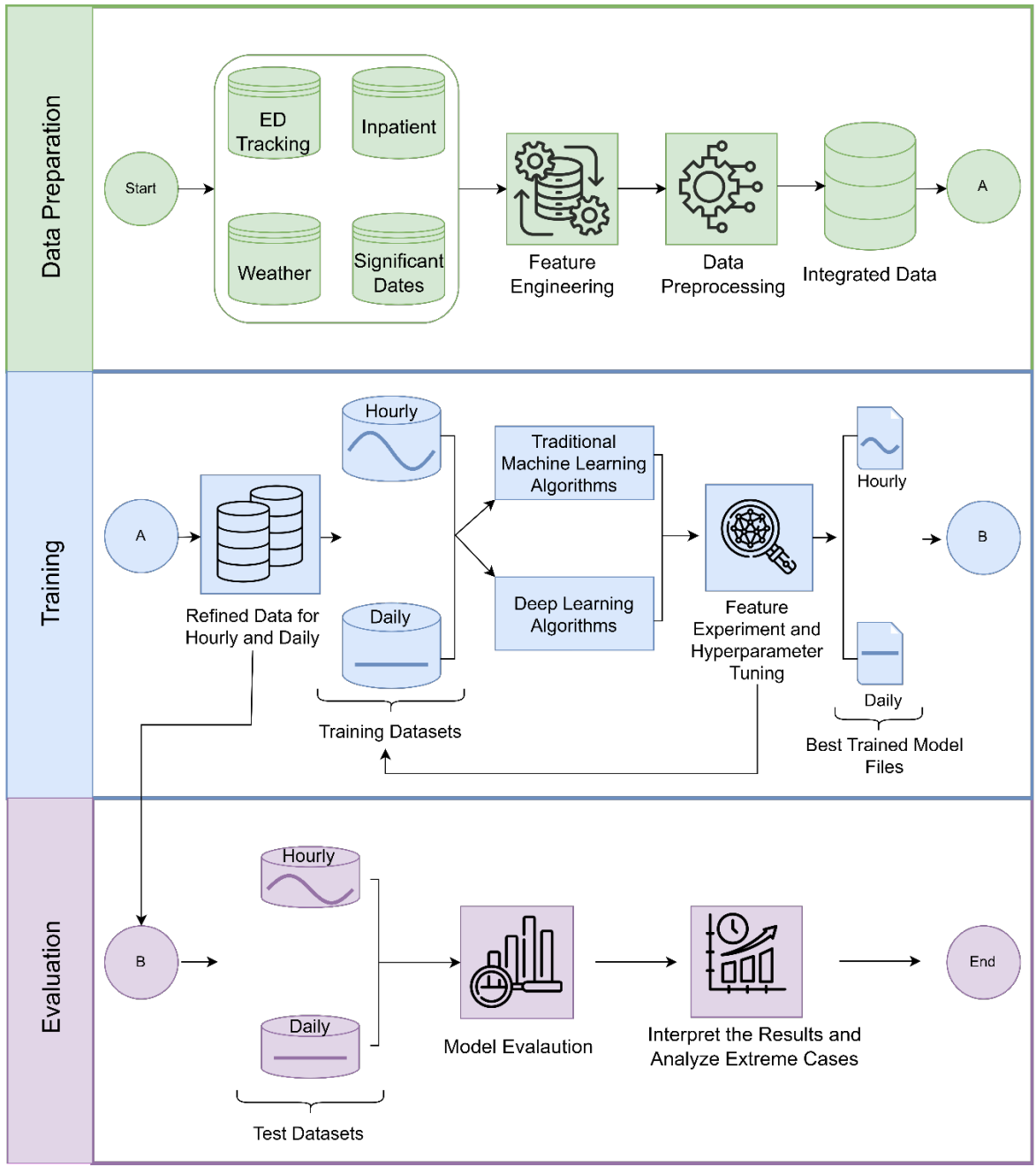

The data preparation phase involves processing data from four sources: the ED tracking system, inpatient records, weather information, and significant dates. After obtaining data from different sources, feature engineering and data preprocessing are applied separately to each source before creating the integrated final data. Following this, data preprocessing focuses on cleaning, scaling, and recategorizing certain categorical variables to prepare datasets for use in predictive modeling. After completing feature engineering and data preprocessing for each data source, all sources are integrated on an hourly basis into a single comprehensive data.

During the training phase, the integrated data from the data preparation stage is used to generate two refined datasets: one for the hourly prediction approach and another for the daily prediction approach. Subsequently, sixteen different dataset variations are created to be used by both approaches, each with distinct feature combinations, as detailed in Table 2. A total of eleven ML algorithms, listed in Table 3, are used to train and evaluate models on both hourly and daily datasets to identify the best-performing models. Each combination of an algorithm and a dataset is considered a separate model in this study. To improve performance, different hyperparameter configurations are tested across the sixteen dataset variations. The best-trained model hyperparameters for each dataset variation and algorithm combination are identified iteratively during training.

In the evaluation phase, performances of the trained models are evaluated. The best-performing models, selected for both hourly and daily prediction tasks, are tested using evaluation metrics outlined in the metrics subsection. To further assess model effectiveness, the best-performing hourly prediction model is evaluated under extreme case scenarios, representing periods of exceptionally high patient volumes, with the results presented in Table 4. Additionally, the performance of the best hourly model is analyzed across the hour-of-day, capturing variations and trends in predictive accuracy at different times, as illustrated in Figure 4.

### Data Sources

The data sources in this study cover the period from January 2019 to July 2023 and are categorized into hospital data, including ED tracking system and inpatient records, and external data, consisting of weather information and significant dates for football games and federal holidays.

The ED tracking data provides detailed records of patient arrivals and departures within the ED waiting and treatment rooms of our partner hospital. This data also includes emergency severity index (ESI) levels indicating acuity, patient classifications, room types, event status (e.g., complete, request, cancel), and other ED-related information. Both waiting room and treatment room identifiers are recorded, enabling comprehensive tracking of patient movements from arrival to discharge or inpatient admission through unique patient and visit identifiers. The data contains 161,477 unique patients and 308,196 unique visits.

The inpatient data contains time-stamped records of patient admissions to and discharges from inpatient units. These records enable the hourly calculation of hospital-wide patient census feature, as shown in Table 1. The dataset comprises 293,716 unique inpatients visits for 180,589 unique patients, representing hospital-wide admissions encompassing the study period.

The external data includes weather data and significant events, which are federal holidays and local football game event data. The weather data provides hourly weather actual information collected from a nearby weather station located close to the partner hospital. This data is sourced from the historical data archive of the OpenWeatherMap API [25] and includes numerical variables such as temperature, humidity, and wind speed, as well as a categorical variable indicating categories of clear skies, clouds, rain, mist, thunderstorm, snow, drizzle, haze, fog, and smoke. The significant dates consist of two external datasets: federal holidays and football game dates for a major team near the hospital. On average, there are 13 football games and 10 federal holidays each year. Federal holidays were obtained from the United States Office of Personnel Management Government website [26], and football game dates were sourced from the team's official website [27].

## Problem Modeling

To predict waiting counts in the ED waiting room at a time point $h$ steps into the future (e.g., predicting six hours from now, or any other chosen future time-point), we model this problem as a timeseries machine learning problem. We set up the problem as a direct single-step forecast targeting $y_{t+h}$ from the current time $t$. Let:

- $y_t$ represents the waiting count at the current time $t$.
- $h$ be the prediction horizon (i.e., the number of time steps ahead for the forecast (e.g., $h=6$ for six hours ahead).
- $k$ denote the number of past observations (lags) included in the model.
- $(l_1, l_2, \ldots, l_N)$ represent variables that do not require lagging (e.g., weather conditions, holidays).
- $x_t, x_{t-1}, \ldots, x_{t-k+1}$ represents any feature other than waiting count that has lag.
- $g(.)$ be the function learned by the machine learning model, trained using historical data.

The model can be expressed as follows:

$$y_{t+h} = g(y_t, y_{t-1}, \ldots, y_{t-k+1}, x_t, x_{t-1}, \ldots, x_{t-k+1}, \ldots, x_t, x_{t-1}, \ldots, x_{t-k+1}\}, l_1, l_2, \ldots l_N), \text{for } t = k+1, k+2, \ldots, M-h \quad (1)$$

Where:
- $y_{t+h}$ is the predicted waiting count at time $t+h$.
- $y_t, y_{t-1}, \ldots, y_{t-k+1}$ are waiting count lag features to capture temporal dependencies.
- $M$ is the total number of observations and the training dataset size is $M - k - h + 1$.

## Feature Engineering and Data Preprocessing

In the context of machine learning, predictors or variables are often referred to as features. We applied a series of feature engineering and preprocessing steps to transform raw data into structured input for ML algorithms. These steps make the data ready for modeling as described in Equation 1. First, each of the four data sources, including ED tracking data, inpatient records, weather data, and significant event dates, were processed separately to extract meaningful features before applying preprocessing steps. All features were calculated on an hourly interval, where each row represents a one-hour time step in a continuous sequence. The feature engineering process is summarized as follows:

- **Date Features:** Temporal information for each row in the training and test data was represented by breaking down each timestamp into individual components, including the calendar year, the numerical month (e.g., January as 1), the specific day within the month (1-31), the day of the week as an integer (e.g., Monday as 0 and Sunday as 6), and the hour of the day in a 24-hour format. The dataset is organized into an hourly frequency, with each row corresponding to a one-hour interval. These date features are common across all data sources.

- **Waiting Count:** The ED tracking dataset was filtered to include only waiting room encounters and grouped by the visit ID. For each patient-visit combination, the earliest arrival $A_i$ and latest departure $D_j$ times in the waiting rooms were identified to track patient presence. The total hourly waiting count, *N(t)*, can be calculated by summing all patients present within each hourly interval starting at the hour *t*, where a patient is counted if their earliest arrival *Ai* occurs before the end of the hour *t+1* and their latest departure $D_j$ is at or after the start of the hour *t*, as represented in Equation 2. The indicator function *I* is equal to 1 if these conditions are met and 0 otherwise, and $n(t)$ represents the total number of patients considered for hour *t* (i.e., the number of unique visits during *[t,t+1]* hourly interval). This hourly waiting count serves as the target feature for our predictive modeling in this paper.

$$N(t) = \sum_{i=1}^{n(t)} I\left(A_i < t + 1 \text{ and } D_j \geq t\right) \qquad (2)$$

- ***Waiting Counts by ESI Levels:*** The waiting count at each hourly interval was also calculated for three categories of patients based on their ESI levels: 1-2 (very urgent), 3 (urgent), and 4-5 (non-urgent). The hourly waiting count for each group was calculated using the same method as the overall hourly waiting count, relying on arrival and departure times to track patient presence for each ESI level category.

- ***Average Waiting Time:*** Waiting times were calculated for each patient who fits the criteria for the waiting count based on their ED arrival and departure times within a specific hourly interval, ensuring an accurate measurement of time spent in the waiting room. The average waiting time for each hourly interval *[t, t+1]*, Average Waiting Time (*t*), was calculated by summing the waiting times $WT_i(t)$ for all waiting patients and dividing by the number of waiting patients $N(t)$ during that hour, as represented in Equation 3.

$$Average\ Waiting\ Time\ (t) = \frac{\sum_{i=1}^{n(t)} WTi(t)}{N(t)} \qquad (3)$$

- ***Average Waiting Times by ESI Levels***: The hourly average waiting time was calculated separately for three patient groups based on their ESI levels. The calculation followed the same approach as the overall average waiting time but was applied within each ESI level category. Specifically, the arrival and departure times of patients within each ESI group were used to determine individual waiting times, which were then averaged within that group.

- ***Treatment Count***: The treatment count for each hourly interval was calculated using the same method as the waiting count. However, instead of considering waiting rooms, only patients in treatment rooms were counted. This was done by filtering the data to include only treatment room locations when tracking patient arrivals and departures.

- ***Average Treatment Time:*** The average treatment time was calculated similarly to the overall average waiting time but focused specifically on patients in treatment rooms.

- ***Extreme Case Indicator:*** The Extreme Case Indicator (ECI) is a binary feature used to identify periods of exceptionally high patient volume. It is assigned a value of 1 when the hourly waiting count ($WCt$) at time *t* exceeds a statistical threshold, calculated as the mean (μ) plus two standard deviations (σ) of the waiting count distribution, as defined in Equation 4. Here, $WCt$ represents the actual number of patients waiting during a specific hourly interval. This threshold accounts for natural variations in patient volume, ensuring that only significantly high values are classified as extreme. If $WCt$ remains below this threshold, the ECI is set to 0, indicating a non-extreme period.

$$ECI(t) = \begin{cases} 1, & if\ WCt \geq \mu + 2\sigma \\ 0, & otherwise \end{cases} \tag{4}$$

- **Boarding Count:** Boarding patients are those who have completed ED treatment, decided to be admitted, and are waiting for an inpatient bed to become available after a bed request has been made. The boarding period begins when the inpatient bed request is submitted and ends when the patient leaves the ED for the inpatient unit (marked by the ED checkout timestamp). Using these start and end times, the hourly boarding count was calculated similar to the waiting counts calculation. For instance, if a patient receives an inpatient bed request at 10:10 AM and leaves the ED for the inpatient unit at 11:15 AM, their boarding period spans from 10:10 AM to 11:15 AM. Since boarding counts are calculated on an hourly basis, this patient is counted in both the 10:00 - 11:00 AM and 11:00 - 12:00 PM intervals.

- **Average Boarding Time:** The boarding time for a particular boarding patient was calculated by subtracting ED checkout time from inpatient bed request time. Subsequently, the average boarding time for each hourly interval was calculated by summing individual boarding times of the patients included in the boarding count for that interval and dividing it by the boarding count.

- **Hospital Census:** This feature represents the total number of inpatients in the hospital for a given hourly interval, excluding patients in the ED. It is calculated hourly to reflect the occupancy of inpatient units.

- **Weather features:** A categorical feature representing atmospheric conditions including clear skies, clouds, rain, mist, thunderstorms, snow, drizzle, haze, fog, and smoke. This feature captures hourly weather patterns, with each hourly interval potentially having a different weather status, represented as categorical variables in the dataset. Temperature, humidity, and wind speed were considered as part of the weather features.

- **Football Game:** This categorical feature indicates whether a football game is occurring during a given hourly interval, based on the schedule of a major team near the hospital.

- **Federal Holidays:** This categorical feature indicates whether a specific day is a federal holiday, such as Christmas or Independence Day. When a federal holiday occurs, the feature is assigned the corresponding label for all 24-hourly intervals of that day in the hourly dataset.

- **Leg generation**: Lag features were created by shifting past values of the target variable to serve as additional input features, allowing the model to capture temporal patterns. For example, a 12-hour lag feature set includes waiting counts recorded at each hour from 1 to 12 hours before the current time (i.e., values at $t-1, t-2, \ldots, t-12$). Similarly, 24-hour and 48-hour lag sets include values from the past 24 and 48 hours, respectively. These lag intervals were selected based on dataset characteristics, as detailed in Table 2.

- **Rolling mean calculations:** These were applied to reduce short-term variability by averaging values over a defined window. For instance, a 4-hour waiting count rolling mean concept computes the average waiting count values over the current and the previous three hourly intervals, helping to smooth sudden fluctuations. 4 and 6 hours rolling mean windows were selected as detailed in Table 2.

After feature engineering, preprocessing steps were applied to prepare the data for model training and testing, as illustrated in Figure 1. These steps include categorizing specific features to simplify analysis and excluding certain data periods and outliers to minimize potential biases. Additionally, standardization techniques were applied to maintain consistency across numerical features. The following describes the performed preprocessing steps in detail:

- Weather data source included various weather conditions, such as clear skies, clouds, rain, mist, thunderstorms, snow, drizzle, haze, fog, and smoke. To address data sparsity and improve model performance, these conditions were grouped into five categories: Clear, Clouds (combining Clouds and Mist), Rain (combining Rain and Drizzle), Thunderstorm, and Others (including less frequent conditions like snow, haze, fog, and smoke). During dataset creation, two versions were generated: one retaining all ten original weather categories and another using the simplified five-category system to reduce dimensionality, as seen in Table 2.

- The ED tracking data source was preprocessed to exclude instances where patients had been waiting in the waiting room for more than 9 hours. This threshold was determined in consultation with the advisory board from the partner hospital's emergency department. These cases were identified as outliers, deemed as potential data entry errors, that could affect the performance of the predictive models and were removed from the data.

- Data from January 2020 to May 2021, referred to in this study as the COVID-19 period, was excluded due to unusual patterns in patient volumes and hospital operations during the pandemic. This period was marked by reduced patient volumes in both hospitals and the ED, as well as altered hospital procedures, resulting in atypical data that could interfere with the ability of ML models to learn general patterns. For example, as shown in Table 1, the Hospital Census, which has an average of 794, shows a significant drop to 409 on some dates during this period.

- Numerical features from all data sources were normalized using various scaling methods, such as Z-score normalization [28] and MinMaxScaler [29].

### Data Integration and Aggregation

The data collected from various sources were processed through feature engineering and data preprocessing, then merged into a single dataset. The new dataset is referred to as "integrated data" in Figure 1, where each row represents an hourly interval at a frequency of 1 hour. From the integrated data, two distinct datasets were created: one for the hourly prediction model, where each row corresponds to an hourly record, and another for the daily prediction model, where each row represents the average values of the preceding 24 hours from the hourly data.

The hourly dataset contains 27,756 records, each representing a unique hourly interval of the study date range. The daily dataset, created by averaging consecutive 24-hour segments, contains 1,155 records, making it 24 times smaller than the hourly dataset. This aggregation reduces short-term fluctuations, leading to a lower standard deviation in the daily predictions, as detailed in Table 1. For instance, the hourly approach's target variable of waiting count has a mean of 18.11 and a standard deviation of 9.77. While the mean remains the same in the daily dataset, the standard deviation decreases to 4.51 due to averaging. Table 1 provides an overview of the extracted features and their descriptive statistics for both the hourly and daily datasets after they were preprocessed.

**Table 1.** Summary of the Overcrowding Dataset.

| Feature | Date Range, Average ± Standard Deviation (Range) for Numerical Features, % for Categorical Features, and Event Counts | |
|---|---|---|
| | Hourly | Daily |
| Date Range[a] | | |
|     Year | 4 years | 4 years |
|     Month | 12 Months | 12 Months |
|     Day of Month | Days 1-31 | Days 1-31 |
|     Day of Week | 7 Days | 7 Days |
|     Hour | 24 Hours | 24 Hours |
| Waiting Count (Target Variable) | 18.11 ± 9.77 (0 – 59) | 18.11 ± 4.51 (7 – 35) |
| Waiting Count by ESI Levels | | |
|     ESI levels 1&2 | 26.38% | 26.38% |
|     ESI level 3 | 58.34% | 58.34% |
|     ESI levels 4&5 | 14.69% | 14.69% |
| Average Waiting Time | 90.98 ± 62.85 (0 – 425) (minute) | 90.98 ± 32.24 (9 – 170) (minute) |
| Average Waiting Time by ESI Levels | | |
|     ESI levels 1&2 | 62.81 ± 69.98 (0 – 538) (minute) | 62.81 ± 31.69 (5 – 162) (minute) |
|     ESI level 3 | 106.9 ± 76.13 (0 – 526) (minute) | 106.9 ± 38.92 (8 – 193) (minute) |
|     ESI levels 4&5 | 56.03 ± 66.79 (0 – 536) (minute) | 56.03 ± 28.53 (8 – 166) (minute) |
| Treatment Count | 68.29 ± 23.69 (9 – 139) | 68.29 ± 22.64 (31 – 123) |
| Average Treatment Time | 52.93 ± 3.11 (28 – 60) (minute) | 52.93 ± 2.07 (46 – 57) (minute) |
| Boarding Count | 46.78 ± 29.60 (3 – 121) | 46.78 ± 29.30 (8 – 115) |
| Average Boarding Time | 54.06 ± 4.45 (12 – 60) (minute) | 54.06 ± 3.15 (42 – 58) (minute) |
| Extreme Case Indicator | 3.1% | 28.77%[b] |
| Hospital Census | 794.23 ± 71.88 (584 – 1017) | 794.23 ± 61.50 (610 – 931) |
| Temperature | 64.06 ± 15.36 (9 – 100) (°F) | 64.06 ± 13.60 (15 – 87) (°F) |
| Wind Speed | 2.26 ± 1.97 (0 – 15) (m/s) | 2.26 ± 1.22 (0 – 7) (m/s) |
| Humidity | 73.38 ± 19.08 (15– 100) (%) | 73.38 ± 11.97 (40– 97) (%) |
| Weather Status | | |
|     Clouds | 57.67% | 57.67% |
|     Clear | 22.89% | 22.89% |
|     Rain | 15.12% | 15.12% |
|     Mist | 2.55% | 2.55% |
|     Thunderstorm | 1.27% | 1.27% |
|     Drizzle | 0.16% | 0.16% |
|     Fog | 0.13% | 0.13% |
|     Haze | 0.12% | 0.12% |
|     Snow | 0.06% | 0.06% |
|     Smoke | 0.03% | 0.03% |
| Football Game | 40 Games | 40 Games |
| Federal Holidays | 34 Days | 34 Days |
| [a]Data from January 1, 2020, to May 1, 2021, was excluded due to COVID-19. | | |
| [b]28.77% of rows have non-zero values, indicating days that experienced extreme patient volumes during at least some portion of the day. | | |

### Model design

Model design involves selecting the relevant feature combinations for prediction. With a total of 33 main features, excluding lag and rolling mean features, determining the optimal feature set is an essential step in improving model performance. Therefore, numerous experiments were conducted initially to refine feature combinations, scaling methods, COVID-19 date exclusion ranges, weather categorizations, and hyperparameter ranges. Based on the outcomes of these experiments, a total of sixteen datasets were generated, as shown in Table 2, each containing different combinations of features. These datasets were designed to investigate how various feature combinations impact the performance of the prediction models. To achieve this, eleven different algorithms, as detailed in Table 3, were applied to train the models on each dataset. The primary objective was to identify the most effective combination of features and algorithms for achieving optimal performance. For both hourly and daily prediction models, a total of 8,800 experiments (11 algorithms × 16 datasets × 50 trials) were conducted separately to evaluate these combinations and determine the best-performing model and feature set.

**Table 2.** Features and Their Combinations for Dataset Variants.

| Data Sources and Scaling | Features | Lags and Rolling Mean | DS0 | DS1 | DS2 | DS3 | DS4 | DS5 | DS6 | DS7 | DS8 | DS9 | DS10 | DS11 | DS12 | DS13 | DS14 | DS15 |
|---|---|---|---|---|---|---|---|---|---|---|---|---|---|---|---|---|---|---|
| ED Tracking | Waiting Count | Lags (W=24) | X | X | X | X | X | X | X | X | X | X | X | X | X | X |  | X |
| | | Lags (W=48) |  |  |  |  |  |  |  |  |  |  |  |  |  |  | X |  |
| | | Rolling Mean (W=4) |  |  |  |  |  |  |  |  |  |  | X | X | X | X | X | X |
| | | Rolling Mean (W=6) |  |  |  |  |  |  |  |  |  | X |  |  |  |  |  |  |
| | Average Waiting Time | No Lags |  |  |  |  |  |  |  | X | X | X | X | X |  | X | X |
| | | Lags (W=24) |  |  |  |  |  |  |  |  |  |  |  |  |  | X |  |  |
| | | Rolling Mean (W=4) |  |  |  |  |  |  |  |  |  |  |  |  |  | X |  |  |
| | Treatment Count | No Lags |  |  |  |  |  |  |  | X | X | X | X | X | X |  | X | X |
| | | Lags (W=24) |  |  |  |  |  |  |  |  |  |  |  |  |  | X |  |  |
| | | Rolling Mean (W=4) |  |  |  |  |  |  |  |  |  |  |  |  |  | X |  |  |
| | Boarding Count | No Lags |  |  |  |  |  |  |  | X | X | X | X | X | X |  | X | X |
| | | Lags (W=24) |  |  |  |  |  |  |  |  |  |  |  |  |  | X |  |  |
| | | Rolling Mean (W=4) |  |  |  |  |  |  |  |  |  |  |  |  |  | X |  |  |
| | Waiting Count by ESI Levels |  |  |  |  |  |  | X | X | X | X | X | X | X | X | X | X | X |
| | Average Waiting Time by ESI Levels |  |  |  |  |  |  |  |  |  | X | X | X | X | X | X | X | X |
| | Average Treatment Time |  |  |  |  |  |  |  |  |  | X | X | X | X | X | X | X | X |
| | Average Boarding Time |  |  |  |  |  |  |  |  |  | X | X | X | X | X | X | X | X |
| | Extreme Case Indicator |  | X | X | X | X | X | X | X | X | X | X | X | X | X | X | X | X |
| | Year, Month, Day of the Month, Day of the Week, Hour |  | X | X | X | X | X | X | X | X | X | X | X | X | X | X | X | X |
| Weather | Temperature | No Lags |  |  |  | X |  |  |  |  |  |  |  |  |  |  |  | X |
| | | Lags (W=24) |  |  |  |  | X |  |  |  |  |  |  |  |  | X |  |  |
| | Wind speed |  |  |  |  | X | X |  |  |  |  |  |  |  |  |  |  | X |
| | Humidity |  |  |  |  | X | X |  |  |  |  |  |  |  |  |  |  | X |
| | Weather Status (5 Categories) |  |  |  |  |  |  | X | X | X | X | X | X | X | X | X | X | X |
| | Weather Status (10 Categories) |  |  |  |  | X | X |  |  |  |  |  |  |  |  |  |  |  |
| Inpatient | Hospital Census | No Lags |  |  | X | X | X | X | X | X | X | X | X |  |  |  |  |  |
| | | Lags (W=24) |  |  |  |  |  |  |  |  |  |  |  | X | X | X | X | X |
| | | Rolling Mean (W=6) |  |  |  |  |  |  |  |  |  |  |  | X |  | X |  |  |
| Significant Dates | Federal Holidays |  |  |  |  |  |  |  | X | X | X | X | X | X | X | X | X | X |
| | Football Game |  |  |  |  |  |  |  |  | X |  |  |  |  |  |  |  | X |
| | COVID-19 | Exclude between 01-01-2020 and 05-01-2021 | X | X | X | X | X | X | X | X | X | X | X | X | X | X | X | X |
| Scaling | MinMax |  | X |  |  |  |  |  |  |  |  |  |  |  |  |  |  |  |
| | Z-score |  |  | X | X | X | X | X | X | X | X | X | X | X | X | X | X | X |

## Hourly and Daily Time Horizons

To address the varying forecasting needs in ED, this study employs two types of predictive models: hourly and daily, as illustrated in Figure 2. These models were developed to handle different prediction horizons, allowing for more comprehensive management of patient flow in the ED. The hourly model predicts waiting count six hours ahead. As shown by the red arcs in Figure 2, this model provides flexibility by not being restricted to specific times of the day. Instead, it generates forecasts for six hours into the future whenever it runs, enabling it to be applied at any time when updated predictions are needed. By focusing on a six-hours into the future, the hourly model provides predictions that assist in managing resources throughout the day.

The daily model predicts the average waiting count for the next 24-hour period, running each day at 5 PM and utilizing data from the previous 24 hours (5 PM the previous day to 5 PM today). At each run, it provides a single estimate for the upcoming 24-hour window (5 PM today to 5 PM tomorrow), offering insights that aid in planning for the next day. Figure 2 shows the daily model's prediction timeline, represented by the orange elements, where forecasts are generated once per day to estimate the average waiting count in the waiting room for the next 24-hour period. Together, these two models form a comprehensive prediction framework, addressing both immediate and day-ahead planning needs.

Figure 2. Illustration of Hourly and Daily Prediction Models for ED Patient Volume Forecasting.

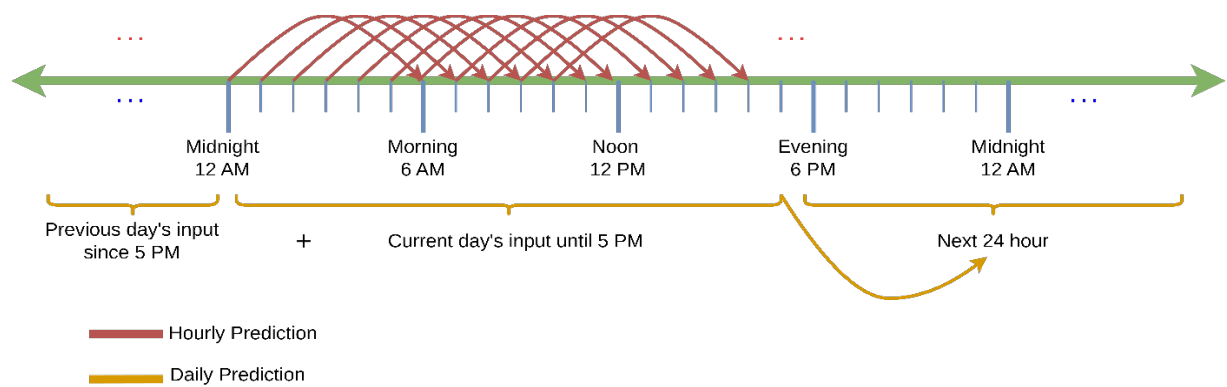

## Model training

We employed eleven ML algorithms to develop the proposed hourly and daily waiting count predictions. The algorithms ranged from traditional algorithms to the state-of-the-art time series algorithms. These algorithms were grouped into four categories: traditional ML methods, RNN-based models, CNN-based models, and transformer-based models. All algorithms were specifically designed to handle time series data. For CNN-based and transformer-based models, we utilized the A State-of-the-Art Deep Learning Library for Time Series and Sequential Data (TSAI) [30], an open-source framework built on PyTorch [31] and Fastai [32], specifically designed for time series tasks. The algorithms were trained using sixteen different datasets, as outlined in Table 2. During the models training, the datasets were divided into 70% training, 15% validation, and 15% testing. Hyperparameter optimization was initially performed using grid search over a predefined subset of the hyperparameter space, tuning key parameters such as learning rate, batch size, dropout rate, weight decay, the number of hidden units, activation function, and other model-specific parameters as applicable.

### Traditional Machine Learning Algorithms

Random Forest (RF) [33] constructs multiple decision trees using bootstrap sampling and random feature selection. By aggregating the outputs of these trees, the model reduces variance, effectively handles high-dimensional data, and minimizes the risk of overfitting. In this study, we used scikit-learn's implementation and experimented with hyperparameters such as the number of trees, tree depth, minimum samples for splits and leaf nodes, and the use of bootstrap sampling (True or False).[34]

Extreme Gradient Boosting (XGBoost) [35] optimizes the loss function using a second-order Taylor expansion, allowing precise gradient approximation and efficient handling of missing data. Regularization techniques reduce overfitting, making it suitable for high-dimensional datasets. In this study, we used XGBoost's implementation and experimented with hyperparameters such as the number of trees, tree depth, learning rate, minimum loss reduction, minimum child weight, tree method, and dropout rate.

### Recurrent Neural Network Based Algorithms

Long Short-Term Memory (LSTM) [36] is a type of RNN designed to handle long-term patterns in sequential data. It uses memory cells with input, forget, and output gates to control what information is retained, updated, or discarded over time. In this implementation, the LSTM model is constructed using the Keras functional API [37], allowing flexibility in architecture design. The model consists of three stacked LSTM layers, each returning both hidden and cell states, followed by dropout layers for regularization. The extracted LSTM features are processed through three dense layers.

Bidirectional Long Short-Term Memory (BiLSTM) [38] extends the LSTM architecture by processing input sequences in both forward and backward directions, enabling the model to effectively capture dependencies from past and future time steps. The model is constructed using the Keras functional API.

Sequence to Sequence Learning with Neural Networks (Seq2Seq) [39] is a neural network framework designed for tasks where input and output sequences differ in length or structure, such as time series forecasting and natural language processing. This model employs an encoder-decoder architecture, where the encoder processes the input sequence and generates a context representation, which is then used by the decoder to reconstruct the output sequence. In this implementation, LSTM layers are utilized within the Seq2Seq framework to effectively capture and utilize temporal dependencies.

### Convolutional Neural Network Based Algorithms

Fully Convolutional Network Plus (FCNPlus) [40] builds on the foundational Fully Convolutional Networks (FCNs), originally developed for semantic segmentation in computer vision [41] and later adapted for time-series modeling [42]. FCNPlus enhances traditional architecture by incorporating residual connections to improve gradient flow and training stability. It also integrates features like dropout for regularization and batch normalization to stabilize and accelerate convergence.

Residual Network Plus (ResNetPlus) [43] builds upon the foundational Residual Network (ResNet) architecture [44], which was originally developed for image recognition in computer vision tasks. ResNet introduced the concept of residual connections to alleviate the vanishing gradient problem and facilitate the training of deep neural networks. In the context of time-series modeling, ResNetPlus adapts this approach by integrating features like separable convolutions, coordinate convolutions, and batch normalization to efficiently capture temporal dependencies and enhance representation learning.

XceptionTimePlus [45] is a specialized deep learning model designed for time series classification, inspired by the principles of the Xception [46] architecture. It extends the foundational ideas of depthwise separable convolutions, as introduced in Xception, to the domain of sequential data. The model is composed of modular XceptionBlocks, which include separable convolutions for efficient feature extraction, bottleneck layers to reduce computational cost, and residual connections to facilitate gradient flow and improve learning stability.

Explainable Convolutional Neural Network for Multivariate Time Series (XCM) [47] is a specialized deep learning architecture tailored for multivariate time series classification tasks. It leverages a hybrid structure combining 2D convolutions to extract inter-variable dependencies and 1D convolutions to capture intra-variable temporal patterns. By integrating these multi-scale features, XCM can effectively model complex temporal relationships across variables. For this study, XCMPlus [48] was used, a variant of XCM with a different backbone structure and feature extraction approach. Unlike XCM, which applies separate 2D and 1D convolutions before concatenation, XCMPlus integrates these layers in a sequential manner. It also incorporates adaptive kernel size selection and adjustments in dropout, batch normalization, and classification layers, modifying the processing flow for multivariate time series classification.

### *Transformer Based Algorithms*

Time Series Transformer Plus (TSTPlus) [49] is a transformer-based model designed for multivariate time series tasks, inspired by Time Series Transformer (TST) [50]. It uses multi-head self-attention to capture both local and global temporal patterns and includes learnable positional encodings to handle sequence structure. The model offers configurable parameters such as attention heads, feedforward dimensions, and encoder layers, making it suitable for various applications.

Time Series Inception Transformer Plus (TSiTPlus) [51] is a time series transformer model that builds on the concepts introduced in the Vision Transformer (ViT) [52]. Inspired by ViT's success in treating image patches as sequential inputs for transformer-based processing, TSiTPlus adapts this idea for time series data. Instead of processing spatial image patches, TSiTPlus treats time series segments as input tokens, enabling it to model long-term dependencies and relationships within sequential data. TSiTPlus processes time series data by dividing it into smaller segments, similar to how Vision Transformers handle image data. The core of TSiTPlus is a self-attention-based encoder, which enables the model to capture dependencies across time steps. The encoder consists of multiple transformer layers, each containing multi-head self-attention [53], feed forward networks, layer normalization, and residual connections. While the model utilizes a multi-head attention mechanism, the exact implementation of the attention module is referenced externally. Additionally, a locality-sensitive self-attention option is available, enhancing attention efficiency for time series patterns. This approach helps the model focus on both individual segments and overall patterns, making it effective for datasets that require understanding long-term trends and dependencies.

**Table 3.** Summary of the machine learning algorithms employed, categorized by their underlying architecture.

| Architecture | Abbreviation | Algorithm |
|---|---|---|
| Traditional Machine Learning Algorithms | RF | Random Forests [33] |
| | XGBoost | Extreme Gradient Boosting [35] |
| Recurrent Neural Network (RNN) Based Algorithms | LSTM | Long Short-Term Memory [36] |
| | BiLSTM | Bidirectional Long Short-Term Memory [38] |
| | Seq2Seq | Sequence to Sequence Learning with Neural Networks [39] |
| Convolutional Neural Network (CNN) Based Algorithms | FCNPlus | Fully Convolutional Network Plus [40] |
| | ResNetPlus | Residual Networks Plus [43] |
| | XceptionTimePlus | Xception Time Plus [45] |
| | XCMPlus | Explainable Convolutional Neural Network Plus [48] |
| Transformer Based Algorithms | TSTPlus | Time Series Transformer Plus [49] |
| | TSiTPlus | Time Series Vision Transformer Plus [51] |

**Model evaluation**

To evaluate the proposed models, they were tested on unseen data. Four standard evaluation metrics were used, which are as follows:

- **Mean Absolute Error (MAE)**: It measures the average absolute difference between predicted and actual values. It provides a straightforward interpretation of prediction error, with lower MAE values indicating better predictive performance.
- **Mean Squared Error (MSE):** It quantifies the average squared differences between predicted and actual values. This metric is sensitive to large errors, making it useful for highlighting significant deviations in predictions.
- **Root Mean Squared Error (RMSE):** It is the square root of MSE, representing prediction errors on the same scale as the target variable. It provides a more interpretable measure of error magnitude while retaining sensitivity to large errors.
- **Coefficient of Determination ($R^2$):** It evaluates the proportion of variance in the target variable that is explained by the model. Values closer to 1 indicate that the model explains most of the variability in the target variable, while a value closer to 0 suggests that the model has little explanatory power.

The models are ranked based on their MAE values, as MAE provides a straightforward interpretation of the average error, making it easier to understand the practical implications of the predictions. Unlike MSE, where improvements can result in larger reductions due to squared differences, MAE decreases more gradually with improvements. Therefore, even small reductions in MAE are significant in this study, as they reflect meaningful enhancements in predictive performance.

To further assess model performance, an Extreme Case Analysis was conducted to evaluate how well the best-performing model predicts rare and critical instances of high patient volume. This analysis considers three levels of severity: Extreme, Very Extreme, and Highly Extreme. An "Extreme Case" corresponds to hourly intervals with waiting counts exceeding 27.88 (mean + 1σ), where σ is the standard deviation. There are a total of 5,097 Extreme Case hourly intervals in the dataset. A "Very Extreme Case" involves waiting counts above 37.65 (mean + 2σ), occurring during 862 hourly intervals, while a "Highly Extreme Case" includes waiting counts greater than 47.42 (mean + 3σ), with 64 such instances in the data. This analysis paves the way for evaluating the model's reliability during rare and critical periods of high patient volume. As detailed in Multimedia Appendix 1, the dataset's descriptive analysis highlights the distribution of waiting counts relative to these thresholds.

To further evaluate the model, its performance was analyzed across each hour of the day over a 24-hour period, providing a detailed view of how prediction accuracy fluctuates throughout the day. This analysis provides insights into the model's performance at different times of the day, correlating with patient volume trends in the ED waiting room. Multimedia Appendix 2 presents the mean and standard deviation of hourly waiting counts, highlighting variations in patient flow. The highest average waiting count occurs at 7 PM, with a mean of 26.23 and a standard deviation of 9.06, reflecting the peak of daily activity. Conversely, the lowest average waiting count is recorded at 6 AM, with a mean of 6.97 and a standard deviation of 4.31, indicating a lull in ED activity during early morning hours.

## Results

We present and analyze the evaluation results of eleven ML algorithms (Table 3) applied to sixteen datasets (Table 2). The results are structured into two parts: first, the performances of models predicting waiting counts six hours ahead are examined, followed by the evaluation of models estimating the average waiting count for the next 24-hour period.

### Performances of Hourly Prediction Models

The target variable waiting count, which represents the number of patients in the waiting room during a given hourly interval, has a mean value of 18.11 and a standard deviation of 9.77 as shown in Table 1. Figure 3 presents the performance analysis of ML models in predicting waiting counts six hours ahead across sixteen different datasets (DS0 to DS15). In Figure 3, the x-axis in all subplots represents the datasets, while the y-axis corresponds to each evaluation metric. Among the evaluated models, TSiTPlus algorithm demonstrated the best overall performance, achieving an MAE of 4.195, an MSE of 29.320, an RMSE of 5.414, and an $R^2$ of 0.56 on dataset DS15.

As shown in Figure 3, traditional ML algorithms, RF and XGB displayed similar performance patterns, with RF achieving its best results on dataset DS15 and XGB on dataset DS14. Specifically, RF's performance on DS15 yielded an MAE of 4.653, an MSE of 34.569, an RMSE of 5.879, and an $R^2$ of 0.486, using 100 estimators, a maximum depth of 30, and 4 samples per leaf, with bootstrapping enabled. Across all datasets, RF's MAE varied between 4.65, its best performance, and 4.786, its worst performance. On the other hand, XGB achieved slightly better values on dataset DS14, with an MAE of 4.62, an MSE of 33.40, an RMSE of 5.78, and an $R^2$ of 0.50, with a maximum depth of 15, a learning rate of 0.02, a subsample ratio of 0.8, and a column sampling rate of 0.3. These results reflect the limitations of traditional algorithms in capturing long-term dependencies and complex patterns naturally present in time-series data. Both methods treat each observation as independent, lacking the capability to account for sequential relationships or long-term dependencies in time-series data.

RNN-based models, including Seq2Seq, LSTM, and BiLSTM, delivered better results than traditional ML algorithms for predicting waiting counts six hours ahead. Among these, Seq2Seq achieved its best performance on dataset DS12, with an MAE of 4.527, an MSE of 34.067, an RMSE of 5.836, and an $R^2$ of 0.493. This result was obtained using a batch size of 32, a learning rate of 0.01, a weight decay of 0.2, a dropout rate of 0.1, and Adam as the optimization function. The Seq2Seq model's performance, based on MAE, ranged from its worst result of 5.28 on dataset DS0 to its best result of 4.52 on dataset DS12. LSTM demonstrated consistent results across datasets, delivering its best performance on dataset DS11, with an MAE of 4.523, an MSE of 33.264, an RMSE of 5.767, and an $R^2$ score of 0.5056. This result was achieved using a batch size of 32, a learning rate of 0.01, a dropout rate of 0.2, a weight decay of 0.1, and Stochastic Gradient Descent (SGD) as the optimization algorithm [54]. The LSTM model's performance, based on MAE, ranged from its worst result of 5.03 on dataset DS0 to its best result of 4.52 on dataset DS11. BiLSTM, which leverages its bidirectional architecture, achieved its best results on dataset DS15, with an MAE of 4.54, an MSE of 34.67, an RMSE of 5.89, and an $R^2$ score of 0.48. However, it encountered higher errors on datasets such as DS2, where the MAE was 4.91. While RNN-based models outperformed traditional algorithms, their performances were still lower than that of CNN-based and transformer-based models.

The CNN-based algorithms, including ResNetPlus, XceptionTimePlus, FCNPlus, and XCMPlus, demonstrated better performance than both traditional ML and RNN-based algorithms across multiple datasets in this study. Among these, ResNetPlus achieved the best overall performance, particularly on dataset DS15, with an MAE of 4.233, an MSE of 29.928, an RMSE of 5.470, and an $R^2$ of 0.5552. XceptionTimePlus achieved its best results on dataset DS15 as well, with an MAE of 4.2740, an MSE of 30.2799, an RMSE of 5.5027, and an $R^2$ of 0.5500. FCNPlus performed best on dataset DS7, with an MAE of 4.3081, an MSE of 30.3208, an RMSE of 5.5064, and an $R^2$ of 0.5491. Finally, XCMPlus delivered its best results on dataset DS7, with an MAE of 4.285, an MSE of 30.2543, an RMSE of 5.500, and an $R^2$ of 0.550. These results suggest that CNN-based models generally provided more accurate predictions than traditional and RNN-based algorithms.

The transformer-based models, TSiTPlus and TSTPlus, delivered better results compared to other algorithm categories in this study, with TSiTPlus being the best-performing model based on MAE. TSiTPlus achieved its best performance on dataset DS15, with an MAE of 4.1953, an MSE of 29.369, an RMSE of 5.419, and an $R^2$ of 0.563. It also performed consistently across other datasets, such as DS10 (MAE of 4.2070) and DS7 (MAE of 4.2072). In comparison, TSTPlus showed good performance but had slightly higher error values. Its best results were on dataset DS15, with an MAE of 4.242, an MSE of 29.528, an RMSE of 5.434, and an $R^2$ of 0.561. DS15 incorporated a range of features, including waiting count lags, rolling averages, patient flow indicators, weather conditions, hospital census data, and significant event markers. Across other datasets, TSTPlus achieved MAE values such as 4.271 on DS7 and 4.275 on DS8. The performance of TSiTPlus indicates that it was the most accurate model in this study for predicting waiting counts. The strong performance of TSiTPlus in this study is likely due to its ability to handle complex patterns in multivariate time series data. The model uses transformer-based architecture, which is well-suited for identifying relationships and dependencies in sequential data. By dividing time series data into smaller segments and treating these segments as input tokens, TSiTPlus effectively captures both short- and long-term patterns. This design allows it to model dependencies across different timeframes, making it particularly effective for datasets with complex temporal relationships.

**Figure 3.** Comparison of Hourly Model Performances Across 16 Datasets and 11 Algorithms Based on Different Evaluation Metrics

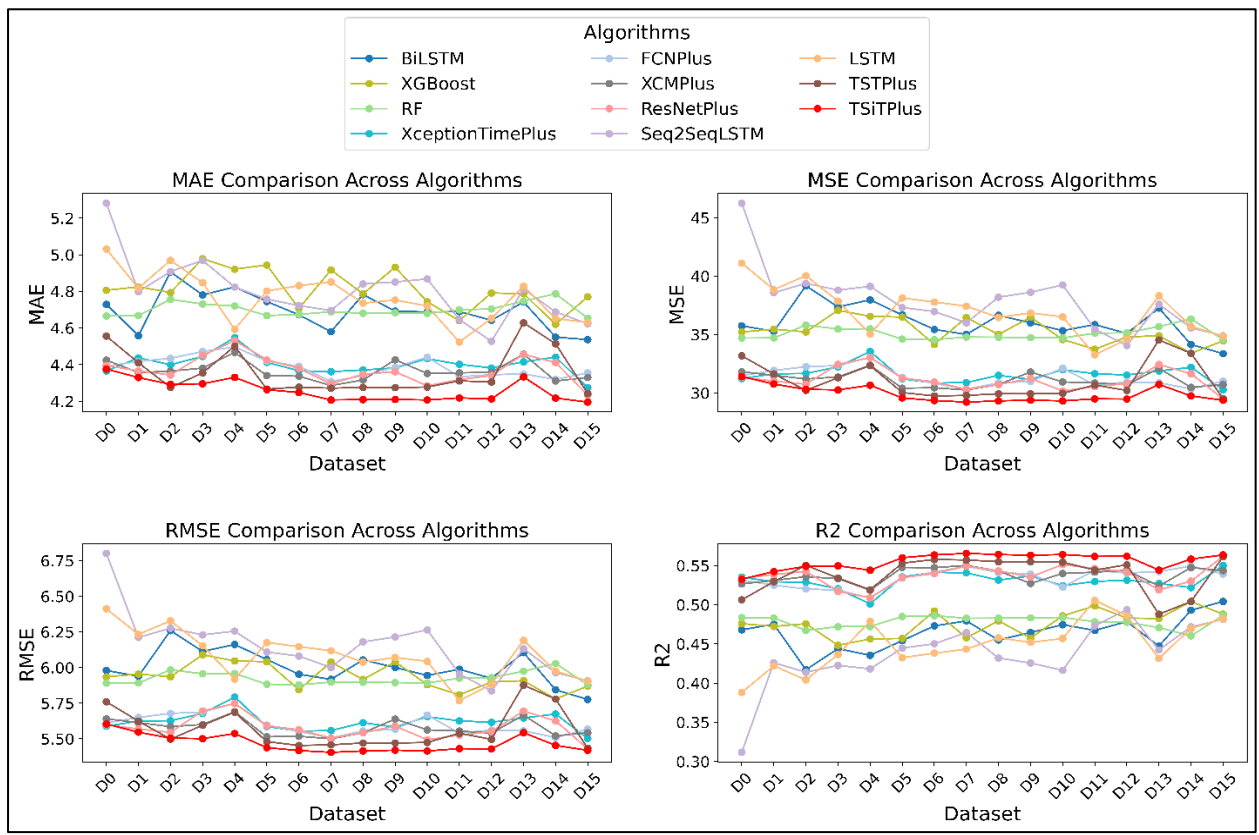

### Analysis of Extreme Case and Hour-of-Day

Table 4 presents the performance of TSiTPlus, the best performing algorithm, in predicting extreme cases across various datasets. For "Extreme Cases" (mean + 1σ), the MAE ranged from 6.16 on dataset DS14 to 7.08 on dataset DS2, with the lowest MSE of 59.34 observed on dataset DS14. For "Very Extreme Cases" (mean + 2σ), the MAE ranged between 10.16 on dataset DS14 and 12.13 on dataset DS1, while the lowest MSE was 126.68 on dataset DS14. In "Highly Extreme Cases" (mean + 3σ), TSiTPlus achieved its best performance with an MAE of 15.59 and an MSE of 261.29 on dataset DS14. Averaging errors across all three extreme case categories, the best performance was observed on dataset DS14, followed by dataset DS15, showcasing their strong predictive capabilities under high patient volume scenarios. This superior performance of DS14 can be attributed to its inclusion of lag features extending to 48 hours, which incorporates information from the past 2 days as shown in Table 2, thereby providing more contextual information for predicting extreme cases.

Our best-performing model not only excels at predicting typical waiting counts but also demonstrates strong predictive capability under extreme crowding conditions. Given that the standard deviation of hourly waiting counts is 9.77, an Extreme Case (≥ 28 patients) represents a scenario where crowding significantly exceeds typical levels. In these situations, the model achieves its MAE of 6.16 on dataset DS14, outperforming all other datasets. For Very Extreme Cases (≥ 38 patients, more than twice the average waiting count), it maintains a low MAE of 10.16, while in Highly Extreme Cases (≥ 48 patients, nearly three times the average), it achieves an MAE of 15.59, highlighting its robustness in handling peak volumes.

**Table 4.** Performance of TSiTPlus in Predicting Extreme Cases Across Different Datasets

| Dataset | Mean + 1σ Extreme (≥ 28) | Mean + 2σ Very Extreme (≥ 38) | Mean + 3σ Highly Extreme (≥ 48) |
|---|---|---|---|
|  | MAE | MAE | MAE |
| DS0 | 6.86 | 11.61 | 17.90 |
| DS1 | 7.07 | 12.13 | 19.30 |
| DS2 | 7.08 | 12.06 | 19.21 |
| DS3 | 6.98 | 11.61 | 18.10 |
| DS4 | 6.85 | 11.36 | 17.77 |
| DS5 | 6.88 | 11.78 | 18.75 |
| DS6 | 6.84 | 11.61 | 18.41 |
| DS7 | 6.63 | 11.21 | 17.54 |
| DS8 | 6.59 | 11.07 | 17.14 |
| DS9 | 6.57 | 11.13 | 17.35 |
| DS10 | 6.58 | 11.11 | 17.15 |
| DS11 | 6.31 | 10.69 | 16.76 |
| DS12 | 6.31 | 10.70 | 16.82 |
| DS13 | 6.18 | 10.44 | 16.78 |
| DS14 | **6.16** | **10.16** | **15.59** |
| DS15 | 6.31 | 10.45 | 16.16 |

Figure 4 illustrates the performance of the TSiTPlus model, highlighting its MAE and RMSE across different hours of the day. The model achieves better performance during nighttime and early morning hours (10 PM to 6 AM), with MAE consistently below 4 and RMSE ranging from 3 to 5.25. These lower error values suggest more stable and predictable patient volumes during these times. In contrast, model performance declines in the late evening hours, particularly between 6 PM and 8 PM, where RMSE exceeds 6 and MAE approaches 5. This pattern aligns with the higher variability in waiting counts observed during these hours, making accurate predictions more challenging.

**Figure 4.** Hour-of-Day Analysis of TSiTPlus Model Performance Using MAE and RMSE

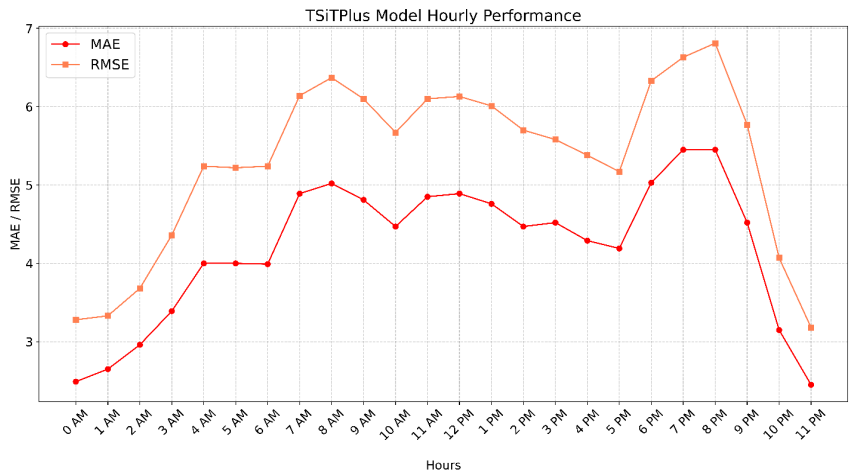

**Performance of Daily Prediction Models**

As shown in Figure 2, the daily prediction is performed at a fixed time each day, specifically at 5 PM. This follows the recommendations of the project advisory board, which suggested building this model to predict the next day's average waiting count. The target variable for daily predictions has a mean value of 18.11 and a standard deviation of 4.51, as presented in Table 1. The prediction error for average waiting count is lower than for hourly predictions due to two factors: first, daily predictions have a reduced standard deviation as they smooth out hourly fluctuations; second, the daily prediction dataset is smaller than the hourly dataset, making it more manageable for the model to process effectively. For the daily predictions, we applied the same eleven ML algorithms listed in Table 3 to sixteen different dataset configurations as detailed in Table 4, mirroring our approach used for hourly predictions. The results of all models are shown in Figure 5 using standard metrics. The best-performing model, XCMPlus, achieved an MAE of 2.00, MSE of 6.64, RMSE of 2.57, and an $R^2$ of 0.44.

Traditional ML algorithms RF and XGBoost demonstrated similar performance across all sixteen dataset configurations for daily predictions, as illustrated in Figure 5. Both algorithms achieved their best results on dataset DS14, with RF slightly outperforming XGBoost. Specifically, RF recorded the lowest MAE of 2.138, MSE of 7.109, RMSE of 2.666, and an $R^2$ of 0.217, while XGBoost followed closely with an MAE of 2.167, MSE of 7.311, RMSE of 2.704, and an $R^2$ of 0.195. The performance gap between the best and worst configurations for both algorithms with RF ranging from 2.138 to 2.416 MAE and XGBoost from 2.167 to 2.498 MAE indicates consistent behavior across different feature combinations. RF achieved its optimal performance with 200 estimators, a maximum depth of 40, minimum samples split of 10, and 8 samples per leaf with bootstrapping enabled. Similarly, XGBoost's best configuration utilized a dart booster with a maximum depth of 12, learning rate of 0.02, subsample ratio of 0.8, column sampling rate of 0.4, and weighted sampling with a rate drop of 0.2.

The analysis of the RNN-based models, including LSTM, BiLSTM, and Seq2Seq, shows varying performance across datasets. The LSTM model performs consistently, with MAE values between 2.06 and 2.60 and RMSE between 2.65 and 3.24, achieving the lowest MAE on dataset DS2 and the highest on dataset DS13. BiLSTM performs slightly better, with MAE ranging from 2.06 to 2.42 and RMSE from 2.65 to 3.07, showing the best performance on dataset DS6 and the highest error on dataset DS12. Seq2Seq provides competitive results, with MAE between 2.08 and 2.48 and RMSE from 2.63 to 3.10, performing best on dataset DS5 and worst on dataset DS13. The $R^2$ values indicate that all models exhibit moderate predictive power, with the highest values around 0.41 and the lowest around 0.11. Among the three models, BiLSTM offers the best balance between low MAE and stable $R^2$ values, while Seq2Seq shows more consistent performance across datasets. The results highlight that dataset characteristics significantly impact model accuracy.

The analysis of CNN-based models, including FCNPlus, ResNetPlus, XceptionTimePlus, and XCMPlus, shows varying performance across datasets. FCNPlus has MAE values from 2.03 to 2.44 and RMSE between 2.58 and 3.01, performing best on dataset DS14 and worst on dataset DS11, with $R^2$ values ranging from 0.26 to 0.23. ResNetPlus shows slightly better performance, with MAE values between 2.05 and 2.40, RMSE from 2.59 to 3.02, and $R^2$ values between 0.25 and 0.22. XceptionTimePlus provides competitive performance, with MAE between 2.10 and 2.33 and RMSE from 2.71 to 2.88, achieving its best results on dataset DS8 and the highest error on dataset DS11. XCMPlus outperforms the other models, with the lowest MAE (2.00 to 2.28) and RMSE (2.57 to 2.92), showing the best performance on dataset DS12 and the highest error on dataset DS0. DS12 incorporated waiting count by ESI levels, treatment and boarding counts, and weather conditions, with rolling means and lag features enhancing trend detection. These features contributed to its strong predictive performance, especially for XCMPlus. It also has the highest $R^2$ values, ranging from 0.44 to 0.28, indicating better predictive accuracy. Overall, XCMPlus demonstrates the best performance among all algorithms, not just CNN-based models, achieving the lowest MAE and highest $R^2$ values, making it the most effective model for predicting daily waiting counts across datasets.

The analysis of the transformer-based models, TSiTPlus and TSTPlus, highlights differences in their performance across datasets. TSiTPlus demonstrates MAE values ranging from 2.08 to 2.17 and RMSE values between 2.63 and 2.80, with the best performance observed on dataset DS5 and the highest error on dataset DS14. The $R^2$ values for TSiTPlus vary from 0.41 to 0.20, indicating moderate predictive accuracy. TSTPlus achieves slightly lower MAE values compared to TSiTPlus, ranging from 2.04 to 2.18, and RMSE values between 2.57 and 2.70. The best results are observed for dataset DS4, while the highest error occurs on dataset DS11. The $R^2$ values range from 0.42 to 0.27, showing slightly better predictive power compared to TSiTPlus. Overall, TSTPlus shows slightly better performance than TSiTPlus in terms of lower MAE and higher $R^2$ values across datasets, making it the more effective transformer-based model for predicting waiting counts. However, both models exhibit similar performance trends, and their effectiveness depends on the characteristics of individual datasets.

**Figure 5.** Comparison of Daily Model Performance Across 16 Datasets Based on Different Evaluation Metrics

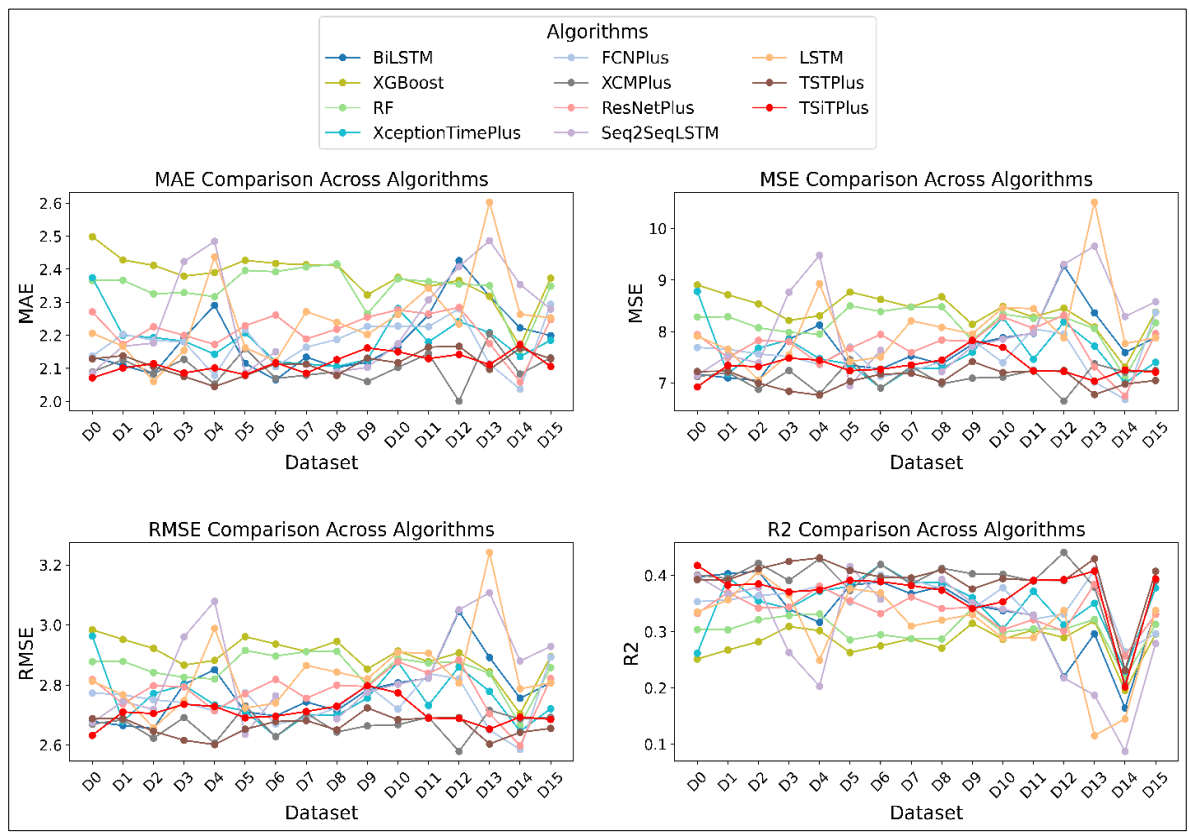

# Discussion

## Managerial Impact

EDs often face overcrowding due to challenges such as limited resources [6], high patient volumes [7], and inefficient patient flow management [8]. This represents an initiative to transform reactive FCP to be proactive by developing predictive models that forecast a key PFM, which is ED waiting counts. The prediction is done at two distinct time scales—six hours ahead and 24 hours in advance—allowing for both hourly operational adjustments and daily planning.

The hourly prediction model, which predicts the total waiting count in the next six hour, enables real-time resource allocation by giving hospital managers enough time (e.g., six hours) to proactively implement FCP interventions resources that help mitigating ED crowding such as adjust staffing, mobilize overflow spaces, and ensure the availability of critical equipment before surges occur. For example, if the model predicts a peak in patient volume during the evening shift, managers can activate a surge plan, offering voluntary short-term shifts to nurses or physicians who opt in to alleviate crowding. Additionally, they can offer overtime to extend staff hours, call in additional clinical staff, or reassign staff to the ED ahead of anticipated crowding. These proactive measures help prevent bottlenecks, ensuring adequate coverage during peak times while minimizing unnecessary overtime costs. This capability minimizes understaffing during peak times while reducing unnecessary overtime costs. Additionally, FCP interventions—such as coordinating with inpatient units to expedite bed turnover or opening additional inpatient surge capacity, such as unstaffed beds or hallways, to absorb some of the ED boarding patients and relieve crowding—can be activated before crowding worsens, improving compliance with key performance indicators such as Centers for Medicare & Medicaid Services door-to-provider times and reducing "left without being seen" (LWBS) rates, which impact hospital revenue and reputation.

The daily prediction model, which estimates the average patient waiting count for the next 24 hours, supports broader decision-making processes. By forecasting average daily patient volumes, managers can optimize next-day staffing schedules and implement strategic interventions such as placing the hospital on diversion to temporarily halt nonessential incoming transfers, thereby preserving ED and inpatient capacity for critical cases. These predictions also enhance cross-departmental collaboration by facilitating proactive intensive care unit (ICU) bed reservations for anticipated ED admissions or diverting non-urgent cases to alternative care settings, such as urgent care clinics, to alleviate ED congestion. Over time, insights from daily predictions inform long-term capacity planning, such as adjusting seasonal budgets for flu surges or expanding ED infrastructure to accommodate growth trends.

The performance of the best model in predicting extreme cases indicates that the model effectively adapts to high-demand scenarios, allowing hospitals to anticipate and respond to surges with greater accuracy. By providing reliable predictions even during severe overcrowding, where waiting counts far exceed typical levels, the model enables better resource allocation, timely patient care, and reduced operational strain, ultimately supporting a more proactive FCP implementation.

**Limitations**

A key limitation of this study is that we did not evaluate how the predictions directly support proactive FCP implementation. While our models can anticipate patient surges, their real-world impact on operational decision-making remains untested. In a future study, we will address this by using discrete event simulation to compare reactive versus proactive FCP strategies, assessing how prediction-driven interventions influence resource allocation, patient flow, and ED performance metrics. Additionally, refining model adaptability and incorporating more external data sources could enhance predictive performance and real-world applicability.

**Conclusion**

This study demonstrates the effectiveness of advanced machine learning models in predicting patient volume in the ED using real-world data from the partner hospital. By integrating multiple datasets—ED tracking, inpatient data, weather, and significant dates—the research developed hourly and daily prediction models. The hourly model provides real-time, six-hour-ahead forecasts, enhancing decision-making and resource allocation. The daily models offer insights into patient volumes at daily averages, aiding in operational planning.

A comprehensive evaluation was conducted by testing eleven different machine learning algorithms on sixteen distinct datasets, each generated through careful feature engineering and hyperparameter optimization to explore optimal feature combinations. The results revealed that for the hourly prediction approach (six-hour ahead prediction), the TSiTPlus algorithm consistently delivered the best performance, with DS15 emerging as the most effective dataset. DS15 incorporated a diverse range of features, including waiting count lags, rolling averages, patient flow indicators, weather conditions, hospital census data, and significant event markers. This combination enabled the model to achieve a MAE of 4.195, MSE of 29.320, RMSE of 5.414, and $R^2$ of 0.56, outperforming all other datasets. For the daily prediction approach (predicting the next 24-hour average waiting count), the best performance was achieved by the XCMPlus model using DS12, with an MAE of 2.00, MSE of 6.64, RMSE of 2.57, and an $R^2$ of 0.44. Through the creation and analysis of multiple feature sets, we identified the most effective feature combinations for improving prediction accuracy, demonstrating the critical role of feature engineering in model optimization.

Furthermore, this study incorporated detailed extreme case and hour-of-the-day analyses to better understand prediction performance under various conditions. The hour-of-the-day analysis further highlighted performance variations throughout the day, providing actionable insights into periods of higher prediction uncertainty and variability. The integration of both hourly and daily prediction models into a real-world decision-support system has significant practical implications for ED management. Such a system can improve resource allocation, optimize staffing, and enhance overall operational efficiency. By leveraging the hourly model's real-time forecasts and the daily model's planning insights, ED administrators can make data-driven decisions to better handle patient flow, reduce overcrowding, and ensure timely care delivery.

## Acknowledgments

This project was supported by the Agency for Healthcare Research and Quality (AHRQ) under grant number 1R21HS029410-01A1.

## Ethical Statement

This study was reviewed and approved by the Institutional Review Board (IRB) at the University of Alabama at Birmingham, with IRB# IRB-300011584.

## Data Availability

The data used in this study is not publicly available due to ongoing research and its confidential nature.

## Conflicts of Interest

None declared

## Abbreviations

**ED$_{(s)}$:** emergency department(s)

**ACEP:** The Emergency Medicine Practice Committee in the American College of Emergency Physicians

**FCP:** full capacity protocol

**PFM$_{(s)}$:** patient flow measure(s)

**PFCT:** flow coordination team

**CPT:** office of clinical practice transformation

**ESI:** emergency severity index

**RNN**: recurrent neural networks

**LSTM:** long Short-Term Memory

**BiLSTM:** bidirectional long short-term memory

**CNN:** convolutional neural networks

**ML:** machine learning

**RF:** random forest

**XGBoost:** extreme gradient boosting

**Seq2Seq:** sequence to sequence learning with neural networks

**FCNPlus**: fully convolutional network plus

**ResNetPlus:** residual network plus

**ResNet:** residual network

**XCM:** explainable convolutional neural network for multivariate time series

**TSTPlus:** time series transformer plus

**TSiTPlus:** time series inception transformer plus

**ViT:** vision transformer

## Multimedia Appendix 1

Distribution of Hourly Waiting Counts with Extreme Thresholds

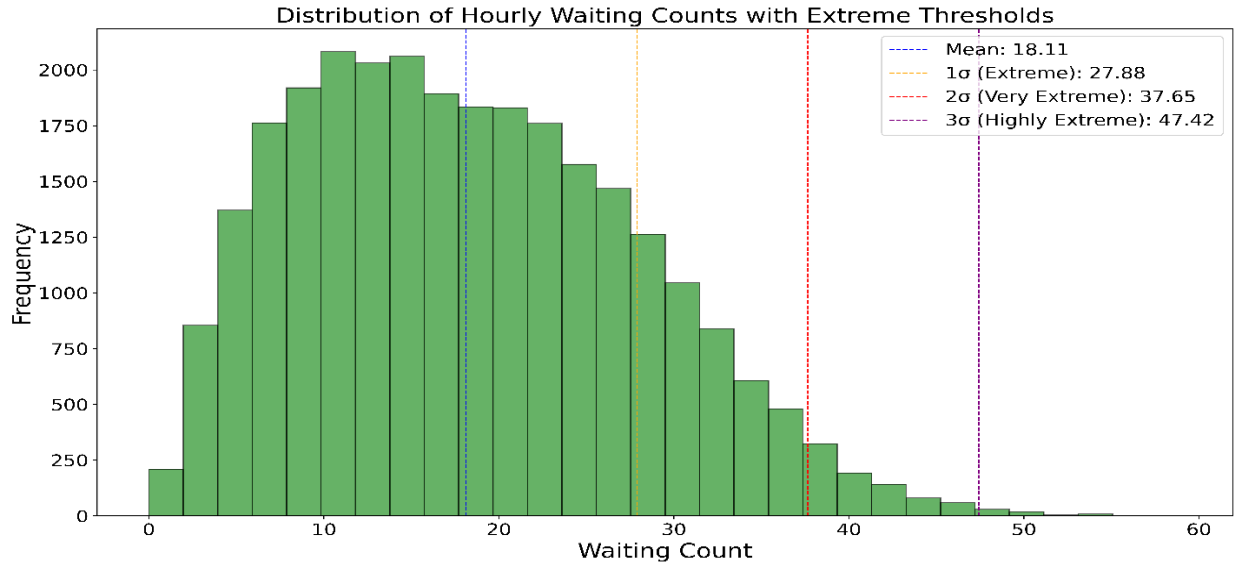

## Multimedia Appendix 2

Mean and standard deviation of hourly waiting counts in the ED waiting room across the Hour-of-Day.

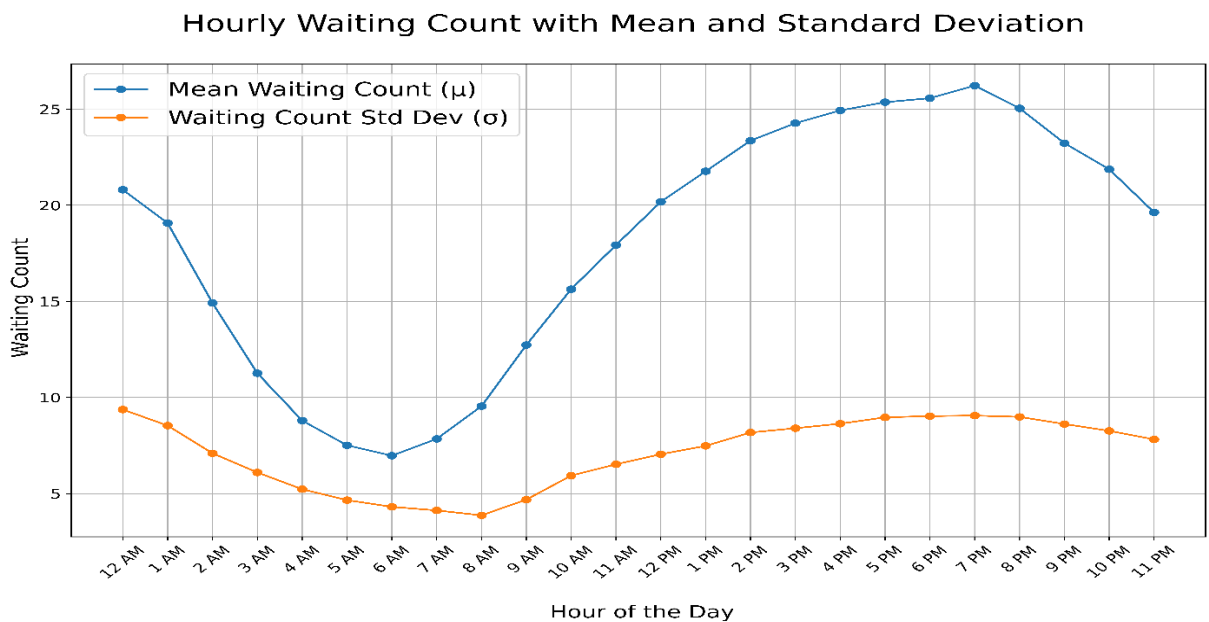